\documentclass[a4paper]{cas-sc}  %,fleqn

\usepackage[numbers]{natbib}
\usepackage{pifont}
\usepackage{amsmath}
\usepackage{caption}
\usepackage{threeparttable}

\def\tsc#1{\csdef{#1}{\textsc{\lowercase{#1}}\xspace}}
\tsc{WGM}
\tsc{QE}
\tsc{EP}
\tsc{PMS}
\tsc{BEC}
\tsc{DE}
\begin{document}
\let\WriteBookmarks\relax
\def\floatpagepagefraction{1}
\def\textpagefraction{.001}

\shorttitle{Cross-Hierarchical Bidirectional Consistency Learning for Fine-Grained Visual Classification}

\shortauthors{P. Gao et~al.}

\title [mode = title]{Cross-Hierarchical Bidirectional Consistency Learning for Fine-Grained Visual Classification}

\author[1]{Pengxiang Gao}[style=chinese]
\ead{pxgao@mail.ustc.edu.cn}

\author[1]{Yihao Liang}[style=chinese]
\ead{liangyihao@mail.ustc.edu.cn}

\author[1]{Yanzhi Song}[style=chinese]
\ead{yanzhis@ustc.edu.cn}
\cormark[1]

\author[1,2]{Zhouwang Yang}[style=chinese]
\ead{yangzw@ustc.edu.cn}

\affiliation[1]{organization={University of Science and Technology of China},
                % addressline={No.96 Jinzhai Road Baohe District}, 
                city={Hefei},
                postcode={230026},
                state={Anhui},
                country={P.R.China}}
\affiliation[2]{organization={Key Laboratory of the Ministry of Education for Mathematical Foundations and Applications of Digital Technology},
                % addressline={No.96 Jinzhai Road Baohe District}, 
                % city={Hefei},
                % postcode={230026},
                % state={Anhui},
                country={P.R.China}}

\cortext[cor1]{Corresponding author}

\begin{abstract}
Fine-Grained Visual Classification (FGVC) aims to categorize closely related subclasses, a task complicated by minimal inter-class differences and significant intra-class variance. 
Existing methods often rely on additional annotations for image classification, overlooking the valuable information embedded in Tree Hierarchies that depict hierarchical label relationships. 
To leverage this knowledge to improve classification accuracy and consistency, we propose a novel Cross-Hierarchical Bidirectional Consistency Learning (CHBC) framework. 
The CHBC framework extracts discriminative features across various hierarchies using a specially designed module to decompose and enhance attention masks and features. 
We employ bidirectional consistency loss to regulate the classification outcomes across different hierarchies, ensuring label prediction consistency and reducing misclassification. Experiments on three widely used FGVC datasets validate the effectiveness of the CHBC framework. Ablation studies further investigate the application strategies of feature enhancement and consistency constraints, underscoring the significant contributions of the proposed modules.
\end{abstract}

% \includegraphics{figs/grabs.pdf}

% \begin{highlights}
% \item Research highlights item 1
% \item Research highlights item 2
% \item Research highlights item 3
% \end{highlights}


\let\printorcid\relax

\maketitle

\section{Introduction}\label{section:I}

\begin{figure}[t]
	\centering
    \includegraphics[width=0.88\linewidth]{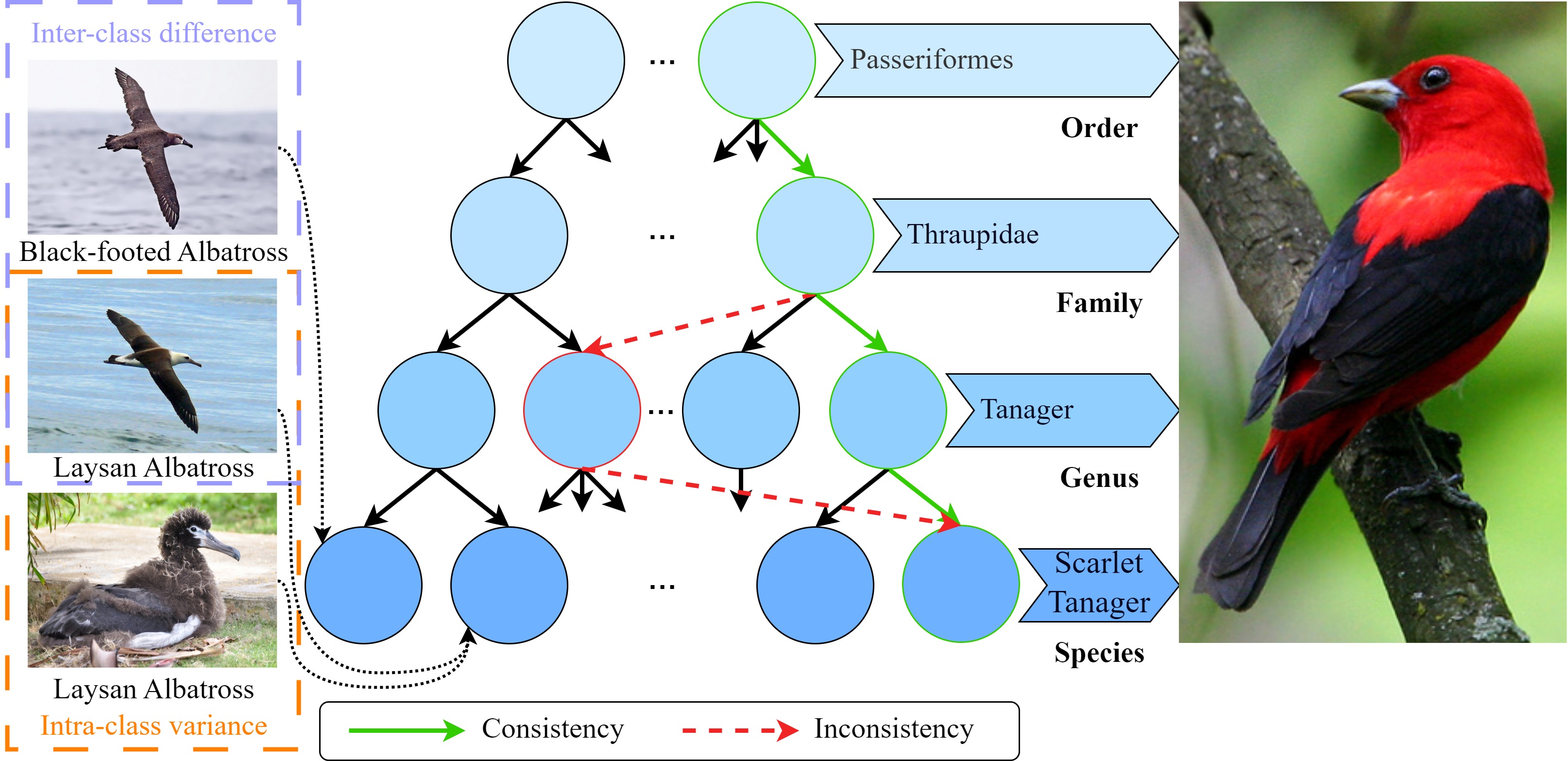}
	\caption{The form of Tree Hierarchy, and we hide the root note. The inter-class difference and intra-class variance in leaf nodes might be limited to subtle aspects like the left. Various users could select the hierarchy they need. Green arrows denote the consistent classification and red arrows mean the inconsistent predictions. }
	\label{FIG:Tree_Hierarchy}
\end{figure}

Fine-grained visual classification (FGVC) aims to categorize images into more specific subclasses within a broader superclass \cite{wang_mpsa_tip24}. This task is challenging due to the substantial intra-class variances and the relatively small inter-class differences at the finer-grained level \cite{yang_p2p_cvpr22}, as illustrated in the left of Figure \ref{FIG:Tree_Hierarchy}. Objects from different subclasses often exhibit subtle distinctions, such as variations in the beak or eye region \cite{wang_neurocomputing25}, while objects within the same subclass can display significant variance in pose, color and other factors, complicating classification. 

Existing methods to address these challenges typically require additional annotations or contextual information \cite{zhao_cvpr21}, with many focusing solely on the finest-grained classification \cite{wang_neurocomputing25, fang_neurocomputing24}. Some strategies use localized annotations \cite{zheng_part_tip20, zhu_dcal_cvpr22, liu_cpcnn_tip22} or additional information \cite{wang_sfff_tcsvt23}, such as latitude data \cite{diao_metaformer_arxiv22}, to enhance accuracy. However, this necessitates extensive labeling for each image, usually performed by domain experts. The need for additional annotations not only increases investment and requires specialized knowledge but also limits model generalization in real-world applications, as these annotations may not always be available.

Moreover, the exclusive focus on the finest-grained classification restricts the model's application scenarios, contradicting the varying granularity required by different users \cite{chang_fgn_cvpr21, wang_cafl_mm23}. For example, a layperson may only need to identify an object as a \textit{bird}, while an ornithologist needs to differentiate specific species, such as the \textit{Scarlet Tanager}. Studies \cite{chang_fgn_cvpr21} show that while experts tend to choose finer labels and amateurs prefer coarser ones, nearly 80\% of expert choices also opt for multi-granularity labels. Classifying in various hierarchies can enhance accessibility and usability for a broader audience \cite{chang_fgn_cvpr21}. However, training a separate model for each hierarchy is laborious and costly \cite{survey}.

Therefore, many approaches focus on training a single model to classify all hierarchies. These methods categorize each level by extracting discriminative features \cite{wang_cafl_mm23, Chen_hrn_cvpr22} or interacting features and results between different levels \cite{liu_chrf_eccv22, chen_hse_mm18}, catering to the diverse needs of various user groups. However, they often overlook the consistency in hierarchical classification \cite{chang_fgn_cvpr21, liu_chrf_eccv22} or underestimate the complementarity of knowledge across different levels \cite{chen_hse_mm18, noor_HD-CapsNet_neurocomputing24}, potentially leading to recognition bias within predicted labels that violate the hierarchical containment relations of their corresponding super- or sub-nodes.

We propose a novel framework, Cross-Hierarchical Bidirectional Consistency Learning (CHBC), which utilizes a Tree Hierarchy  to constrain the consistency of hierarchical classification. The Tree Hierarchy represents the containing relations, with leaf nodes representing the finest-grained classes and non-leaf nodes corresponding to superclasses. This hierarchy is often embedded within the labeling system and aligns with human recognition habits. For instance, if an image is identified as a \textit{Tanager} at a coarse level, it is more likely that its fine-grained label corresponds to a subclass of \textit{Tanager} rather than a subclass of \textit{Albatross}. These containing relations provide a basis for improving classification consistency at different hierarchies and can be easily obtained from the Tree Hierarchy, requiring no additional expert input. Our approach leverages the Tree Hierarchy to enhance accuracy and consistency at each granularity.

The CHBC framework is composed of trunk net, branch net, classifiers, and loss. The trunk net extracts shared image features, while the branch net, consisting of several independent Multi Granularity Enhancement (MGE) modules, amplifies specialized features for each level. Each MGE module is divided into two submodules to generate attention masks and features. The framework predicts labels at various levels using classifiers. A Cross-hierarchical Bidirectional Consistency (CBC) module is designed to calculate classification consistency with the prediction results of each level. This module treats classification results as probability distributions and promotes consistency by minimizing the distance of distributions between different hierarchies. The cross-entropy loss and consistency loss are combined to optimize the CHBC classification results. The CHBC framework offers several advantages: it requires no additional annotations, provides hierarchical labels to meet audience needs, and ensures bidirectional consistency in classifications.

Our main contributions are as follows:
\begin{itemize}
\item We propose a CHBC framework that leverages the inherent hierarchical relationship of labels, eliminating the need for additional labeling information and achieving a high-accuracy classification model. 
\item We introduce a MGE module that enhances the consistency of discriminative spatial information by decomposing multi-granularity features and attention masks. 
\item We design a CBC module that unifies the prediction of hierarchical labels from coarse to fine and vice versa, thereby improving classification consistency across different hierarchical levels. 
\end{itemize}

The remainder of this paper is organized as follows: Section \ref{section:II} reviews related work; Section \ref{section:III} details the proposed framework; Section \ref{section:IV} presents the experiments and comprehensive analysis; and Section \ref{section:V} concludes our work.

\section{Related work}\label{section:II}

This section briefly reviews the FGVC approaches, which are divided into two categories: single-label-based approaches in Section \ref{section:2.1} and multi-label-based approaches in Section \ref{section:2.2}.

\subsection{Single-label-based approaches}\label{section:2.1}
Single-label-based approaches aim solely to categorize the finest-grained labels. These approaches typically employ feature pyramid network (FPN) \cite{fpn}, part localization or other information to improve the finest classification accuracy. 

The fundamental concept of FPN is to extract features at various scales, where features at larger scales retain spatial details and at smaller scales capture rich semantic information \cite{ding_apcnn_tip21}. 
MPSA \cite{wang_mpsa_tip24} extracts the features of parts on different scales and samples the implicit semantic parts for multi-granularity local retrospective fusion. 
Cross-X \cite{luo_crossx_iccv19} learns multi-scale features and enforces regularization of classification results across multi-scale features to improve robustness. 
PMG \cite{du_pmg_eccv20} progressively integrates multi-granularity features and utilizes a simple jigsaw puzzle generator to form images containing information of different granularity levels. 
However, these features are exclusively designed for at the finest-grained label level, without considering the semantic information provided by hierarchical labels.

Part localization employs anchor boxes to extract local information from images, using discriminative parts identified by boxes to improve fine-grained classification \cite{zheng_pacnn_tip20, zhuang_apinet_aaai20}. 
DCAL \cite{zhu_dcal_cvpr22} captures subtle details between global features and high-response local features, and utilizes negative image samples to enhance the robustness. 
CP-CNN \cite{liu_cpcnn_tip22} uses a context transformer to encourage joint feature learning across different parts under the guidance of a navigator. 
Some models incorporate FPN and part localization. AP-CNN \cite{ding_apcnn_tip21} improves accuracy by enhancing multi-scale representations and localizing discriminative regions through the integration of low-level information. 
CSQA-Net \cite{xu_csqa_arxiv24} models the spatial contextual relation between part descriptors and global semantics, using FPN to extract multi-scale part information and capture more distinct details. 
However, anchor boxes typically necessitate manual annotation of box positions for additional supervised learning. Due to their rectangular shape, anchor boxes also introduce irrelevant background information when applied to irregular parts of objects.

Some approaches improve model performance with additional information. 
SAC \cite{do_sac_cai24} introduces a self-assessment classifier to resolve the ambiguity with top-k prediction classes. 
MetaFormer \cite{diao_metaformer_arxiv22} unites images and various types of meta information (date, location, etc.) for joint learning to improve the finest-grained accuracy. 
SFFF \cite{wang_sfff_tcsvt23} captures heterogeneous and complementary features through spatial-frequency collaboration, constrains spatial-frequency features and facilitates deep integration of the model. But additional information increases the requirement of data.

These single-label-based approaches rely on supplementary annotations to categorize only the finest-grained labels and neglect semantic information and inherent knowledge embedded within Tree Hierarchy, requiring more annotations and limiting their application scenarios. To address these limitations, this study prioritizes multi-label-based approaches, which enhance practical utility by systematically integrating hierarchical labels.

\subsection{Multi-label-based approaches}\label{section:2.2}
Multi-label-based approaches adaptively deploy branches aligned with the depth of Tree Hierarchy and categorize for each level, such as classification for birds at four levels: order, family, genus and species \cite{chen_hse_mm18}. These approaches typically employ a trunk network to extract shared features, while branches refine shared features into level-specific representations for classification at the corresponding granularity. These approaches explicitly explore discriminative patterns at each granularity and consider the interaction of information across different hierarchies. 
CHRF \cite{liu_chrf_eccv22} incorporates predict branches to decompose features and fuse cross-hierarchical orthogonal representations. It also includes an orthogonal region regularization mechanism that clusters the same orthogonal features. 
Based on the level-specific features extracted by branches, CAFL \cite{wang_cafl_mm23} introduces SCAD module, which disentangles features corresponding to different regions and channels. Moreover, it leverages the Tree Hierarchy structure to ensure consistency across predictions at different levels. 
HRN \cite{Chen_hrn_cvpr22} adds coarse-grained features as residual connections to fine-grained features and maximizes the marginal probability of observed ground truth by aggregating information from related labels defined in Tree Hierarchy. 
HCSL \cite{liu_hcsl_pkdd24} uses graph representation learning to integrate hierarchical structural information into the framework and controls the negative impact from coarse to fine by hierarchical knowledge-based validity masks. 
HSE \cite{chen_hse_mm18} employs coarser prediction scores as a prior guide to derive finer representations via a soft attention mechanism and exploits the semantic correlations inherent in hierarchical structure, using the prediction results from the preceding level to regularize the current level.

These multi-label-based approaches largely lack reasonable exploration of consistency and restrict interactions to pairwise adjacent levels. To address this gap, CHBC mines granularity-specific features and quantifies classification consistency bidirectionally. By systematically modeling dependencies and facilitating cross-hierarchical knowledge propagation, CHBC achieves comprehensive semantic integration without requiring supplementary annotations.

\section{Method}\label{section:III}

This section demonstrates the overview in Section \ref{section:3.1}, and elaborates the proposed modules MGE and CBC in Section \ref{section:3.2} and \ref{section:3.3}, which improve classification accuracy from the perspectives of feature enhancement and classification consistency respectively. 

\subsection{Overview}\label{section:3.1}

\begin{figure*}[t]
	\centering
	\includegraphics[width=\textwidth]{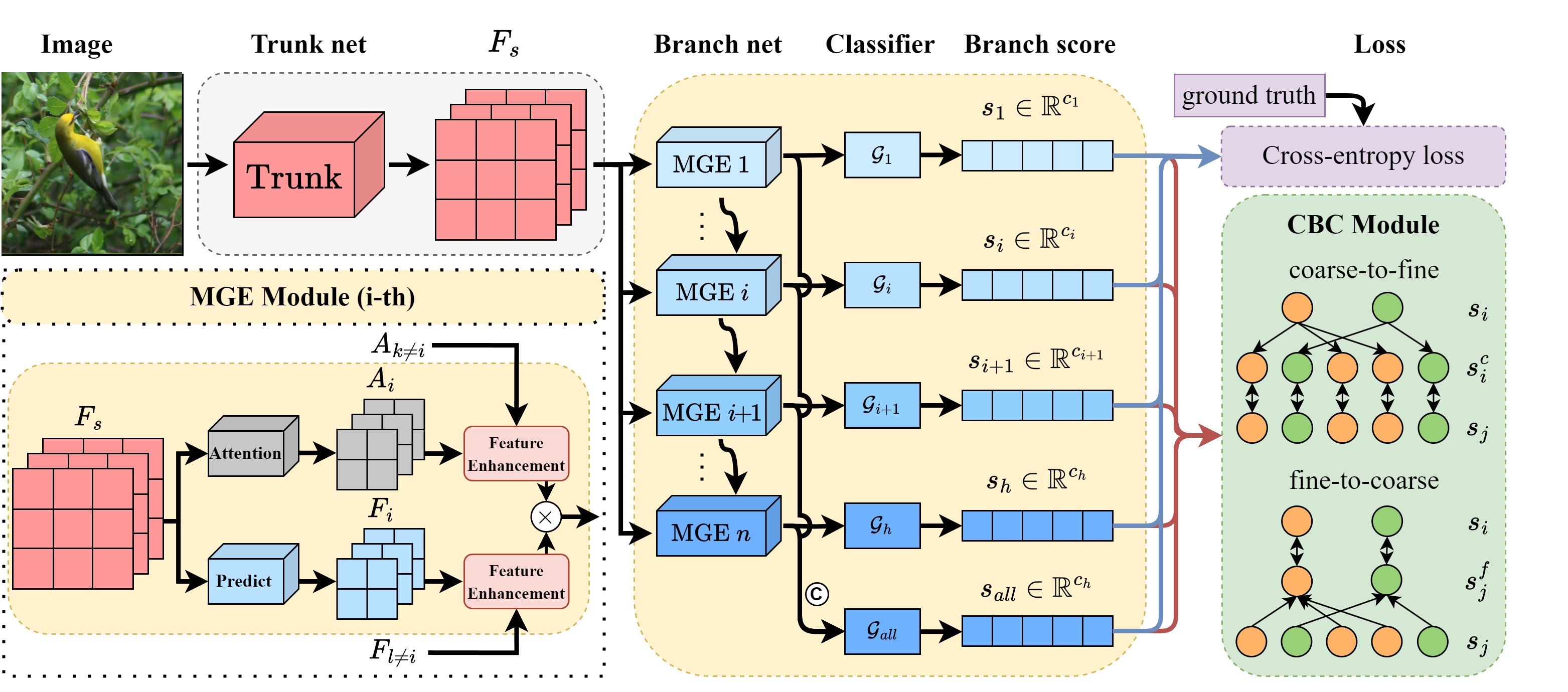}
	\caption{An overview of CHBC. Trunk net extracts shared features. MGE modules in branch net refine and interact features and masks between hierarchies, the bottom left shows the inside of MGE. Classifiers predict hierarchical labels. The right part illustrates two loss functions in CHBC: the consistency loss calculated by CBC and the cross-entropy loss.}
	\label{FIG:framework}
\end{figure*}

The architecture of CHBC is detailed in Figure \ref{FIG:framework}, noting that $h$ is the depth of Tree Hierarchy. Firstly, the trunk net extracts shared features. 
Subsequently, these features are processed by $h$ parallel MGE modules in the branch net, each corresponding to a hierarchical level. 
Each MGE contains two submodules to generate attention masks and features for a specific granularity, respectively. Through feature decomposition across levels, CHBC enhances distinctive representations. Then the enhanced features are fed into classifiers $\mathcal{G}_i, i \in [1, ..., h, all]$, and the labels of $i$-th level are categorized by $\mathcal{G}_i$ and optimized via cross-entropy loss. Ultimately, all hierarchical predictions are evaluated by the CBC module to calculate the classification consistency loss as a part of total loss. 

\subsection{Multi-granularity enhancement}\label{section:3.2}

Many methods employ attention masks to amplify discriminative features by guiding models focus toward salient regions, with higher attention scores indicating more significant regions. Class Activation Mapping (CAM) \cite{cam} provides effective and simple visualization of model attention and can also be served as a spatial attention mechanism. We adapt CAM to better suit CHBC's requirements, integrating it as the spatial attention component within MGEs.

In the branch net, we set up a separate MGE for each granularity to mine the distinct features and decompose information across levels. Each MGE contains two parallel submodules: attention submodule generating attention masks and predict submodule extracting level-specific features from shared features. %and combines them with the attention masks for enhancement in current level. 
Given an image $I$ with hierarchical labels $\{y_1, ..., y_h\}$ from coarse to fine, the trunk net and branch net extract features as below: 
\begin{equation}\label{eq1}
    F_s = \phi_t(I) \ , % \in \mathbb{R}^{C_1 \times H_1 \times W_1}, 
\end{equation}
\begin{equation}\label{eq2}
    A_i = {\rm CAM}( \phi_{a_i}(F_s) ) \ , % \in \mathbb{R}^{C_2 \times H_2 \times W_2}, 
\end{equation}
\begin{equation}\label{eq3}
    F_i = \phi_{p_i}(F_s) \ , % \in \mathbb{R}^{C_2 \times H_2 \times W_2}, 
\end{equation}
where $\phi_t(\cdot)$ is the trunk net, $F_s \in \mathbb{R}^{C \times H \times W}$ represents the shared features, where $C$, $H$, $W$ denote the number of channels, the height and width of features. In the $i$-th MGE ($i \in [1, 2, ..., h]$), $F_s$ first passes through the attention submodule $\phi_{a_i}(\cdot)$ and is classified to obtain attention masks $A_i$ by $\rm CAM(\cdot)$, and the specific features of $i$-th level $F_i$ are extracted by the predict submodule $\phi_{p_i}(\cdot)$. 
Matrix orthogonal decomposition is designed to enhance the distinct information between hierarchies, as illustrated in Figure \ref{FIG:details_in_module} (a). The orthogonal decomposition as Eq. (\ref{eq4}) will be performed both for attention masks and features, which allows CHBC to distinguish high-response local features. 
\begin{equation}\label{eq4}
    M_{orth} = {\rm MOD}(M_{fine}, M_{coarse}) \ , 
\end{equation}
% \begin{align}\label{eq4.1}
%     {\rm MOD}(& M_{fine}, M_{coarse}) = 
%     \notag
%     \\& M_{fine} - \frac{\sum_{(m, n)} (M_{fine}^{(m, n)} \cdot M_{coarse}^{(m, n)}) }{\sum_{(m, n)} (M_{coarse}^{(m, n)} \cdot M_{coarse}^{(m, n)}) } M_{coarse} \ , 
% \end{align}
\begin{align}\label{eq4.1}
    {\rm MOD}(M_{fine}, M_{coarse}) = 
    M_{fine} - \frac{\sum_{(m, n)} (M_{fine}^{(m, n)} \cdot M_{coarse}^{(m, n)}) }{\sum_{(m, n)} (M_{coarse}^{(m, n)} \cdot M_{coarse}^{(m, n)}) } M_{coarse} \ , 
\end{align}
\begin{equation}\label{eq5}
    M'_{fine} = M_{fine} + \alpha \cdot M_{orth} \ , 
\end{equation}
$M_{fine}$ and $M_{coarse}$ represent fine-grained and coarse-grained matrices (feature or attention), respectively. ${\rm MOD(\cdot)}$ denotes the function that applies matrix orthogonal decomposition and $(m, n)$ means spatial location. $M'_{fine}$ is fine matrix enhanced by orthogonal matrix $M_{orth}$. $\alpha$ is a hyperparameter. 
For attention masks, we utilize the previous level to decompose the current level. In the $i$-th MGE ($i \ge 2$), attention masks $A_{i-1}$ and $A_i$ are fed into Eq. (\ref{eq4}), (\ref{eq5}): 
\begin{equation}\label{eq6}
    A'_{i} = A_{i} + \alpha \cdot {\rm MOD}(A_i, A_{i-1}) \ . % \in \mathbb{R}^{C_2 \times H_2 \times W_2}. 
\end{equation}
The matrix orthogonalization is equal to the orthogonalization of vector pooling from matrix, but spatial information can be reserved. % in the orthogonalization of matrix. 
For the attention masks and features of the first level, they are employed to decompose subsequent levels since there are no coarser-grained ones to enhance them.

To improve the model diversity and differentiate features from attention masks, we leverage the coarsest-grained features to decompose the remaining levels, as the coarsest-grained features contain the richest semantic information. 
We calculate the specific representations of current level and enhance them, then the attention masks are incorporated into features as Eq. (\ref{eq7}): 
\begin{equation}\label{eq8}
    F'_{i} = F_{i} + \alpha \cdot {\rm MOD}({F_i}, {F_1}) \ , % \in \mathbb{R}^{C_2 \times H_2 \times W_2}. 
\end{equation}
\begin{equation}\label{eq7}
    \hat F_i = F'_i \cdot A'_i \ . % \in \mathbb{R}^{C_2 \times H_2 \times W_2}, 
\end{equation}

Since coarse-grained features could compensate for the finest level \cite{Chen_hrn_cvpr22}, an additional classification header $\mathcal{G}_{all}$ is implemented to classify the finest level, which performs no transformations and exclusively concatenates the representations of all levels. 
After obtaining the enhanced features of all levels and pooling them, the average pooling representations are concatenated as Eq. (\ref{eq9}) and fed into $\mathcal{G}_{all}$, which has an input dimension: $C * h$, and an output dimension corresponding to the number of the finest-grained categories. Subsequently, the features $F^p_{i}$ are categorized by classifiers $\mathcal{G}_i$, $i \in [1, ..., h, all]$. % to be the finest-grained classifier. 
\begin{equation}\label{eq9}
    F_i^p = {\rm AvgPool} (\hat F_i) \in \mathbb{R}^{C} \ , \ i=1, 2, ..., h \ , 
\end{equation}
\begin{equation}\label{eq10}
    F_{all}^p = {\rm concat} (F_1^p, F_2^p, ..., F_h^p) \in \mathbb{R}^{C * h} \ , 
\end{equation}
where $\rm AvgPool(\cdot)$ represents average pooling operation and $\rm concat(\cdot)$ denotes concatenate operation.

\subsection{Cross-hierarchical bidirectional consistency}\label{section:3.3}

An indistinguishable image is difficult to determine its fine-grained label. %, e.g., an image is classified as \textit{Procellariiformes} at the order level, but classified as \textit{Scarlet Tanager} at the species level (\textit{Scarlet Tanager} is a subclass of \textit{Passeriformes} rather than \textit{Procellariiformes}). 
However, if its coarse label can be identified with high certainty, it is relatively easy to categorize this image at the fine level with prior knowledge of the coarse level \cite{wang_cafl_mm23}. CBC aims to improve the classification consistency across hierarchies and to assist in categorizing with the prior knowledge. CBC draws on human thinking: if an image is \textit{Tanager}, there is a higher probability that its fine-grained label is a subclass of \textit{Tanager}. Despite the occurrence of misclassification, both wrong predicted label and ground truth should belong to the same superclass.

\begin{figure}[t]
	\centering
    \includegraphics[width=0.45\linewidth]{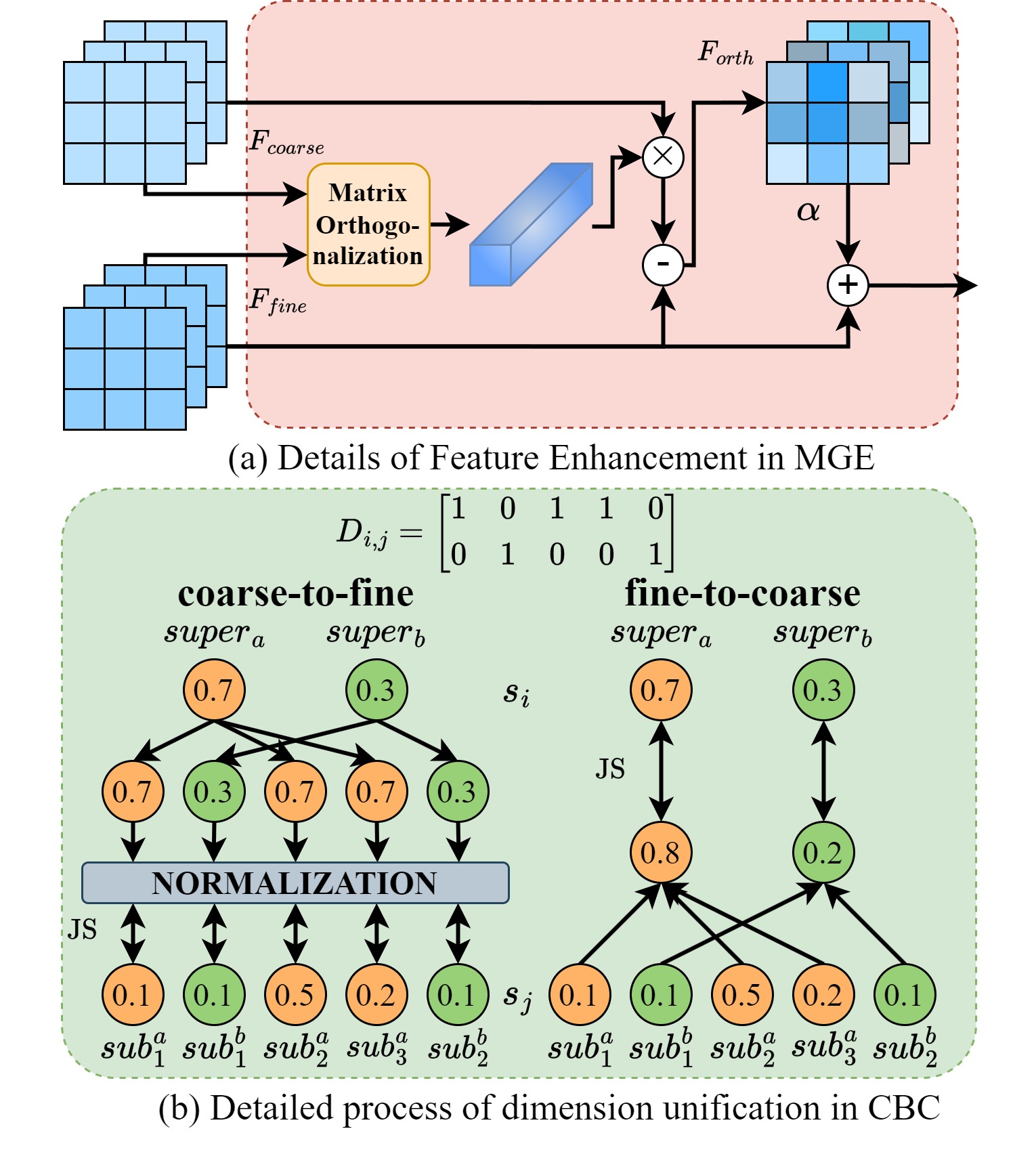}
	\caption{(a) demonstrates the details of feature enhancement in MGE. (b) shows how CBC unifies the distributions of different dimensions into the same dimension, where $sub^a_{[1,2,3]}$ and $sub^b_{[1,2]}$ are subclasses of $super_a$ and $super_b$, respectively.}
	\label{FIG:details_in_module}
\end{figure}

Specifically, multi-level prediction vectors are generated after classifiers, and each vector represents the classification result in a single hierarchy. CBC regards these vectors as probability distributions and minimizes the distance between distributions of different levels. 
As shown in Figure \ref{FIG:details_in_module} (b), for the coarse-grained probability distribution $s_i \in \mathbb{R}^{c_i}$, the dimension $c_i$ indicates the number of categories in the $i$-th hierarchy, while the fine-grained probability distribution is $s_{j} \in \mathbb{R}^{c_{j}}$. CBC first constructs containing relation by Tree Hierarchy. 

Tree Hierarchy could be treated as directed acyclic graph $G(V, E)$. $V = \{ v^1_1, ..., v^1_{c_1}, v^2_1, ..., v^h_{c_h} \}$ denotes the node set and $v^i_e$ represents the $e$-th ($e \in [1, ..., c_i]$) node of the $i$-th hierarchy. $E$ is directed edge set. Adjacency matrix $D$ is the matrix form of $E$. $D$ usually has a large dimension $\sum_{i=1}^h c_i$ of row and column and the binary value in $D$ means the edge between two nodes exists ($1$) or not ($0$). Every node has edges only with its super- and sub-nodes in Tree Hierarchy, leading to the sparsity of $D$. We split $D$ into $h-1$ small matrices $D_{i,i+1}$ to avoid the sparsity and to facilitate dimension unification. $D_{i, i+1} \in \mathbb{R}^{c_i \times c_{i+1}}$ represents the adjacency between the $i$-th and $i\!+\!1$-th level, and $D_{i,j}$ between any two levels ($i<j$) is calculated as: 
\begin{equation}
    D_{i, j} = \prod_{k=i}^{j-1} D_{k,k+1} \ . 
\end{equation}

CBC calculates classification consistency in two directions: coarse-to-fine and fine-to-coarse, and unifies the vectors of different levels to the same dimension by $D_{i, j}$. Coarse-to-fine refers to expanding the coarse-grained distribution $s_i$ to the fine-grained dimension $c_{j}$. 
\begin{equation}\label{eq:coarse2fine}
    s^{c}_{i}=\frac{ s_{i} \times D_{i, j} }{ \sum_{n=1}^{c_{j}} { (s_{i} \times D_{i, j}) [n]} } \ , 
\end{equation}
$s^{c}_{i} \in \mathbb{R}^{c_j}$ is the reflection of classification from coarse to fine, normalizing to ensure the sum of values in $s^{c}_{i}$ equals to 1. 
If the score of superclass $super_a$ is the highest, the scores of its subclasses should also be higher, and the distance between $s^{c}_{i}$ and $s_{i+1}$ should be minimized. 
Fine-to-coarse means that the fine-grained distribution $s_j \in \mathbb{R}^{c_j}$ is congregated to the coarse-grained dimension $c_i$. 
\begin{equation}
    s^{f}_{j} = s_{j} \times D_{i, j}^T  \ ,
\end{equation}
$s^{f}_{j}$ represents the prediction scores that the subclasses belonging to the same superclass are congregated at the coarse hierarchy, and if the subclass $sub^a_{2}$ has a high score, its corresponding superclass $super_a$ should also have a relatively high prediction confidence.

Kullback-Leibler (KL) divergence is the most widely used to measure the distance between probability distributions. 
However, KL requires one distribution to be defined as a reference and another to align with it, which is impractical in experiments. It is unreasonable to determine the coarse or fine distribution as reference. 
Jensen-Shannon (JS) divergence is a variant of KL divergence, leveraging the average of two distributions as a reference and pulling them toward this midpoint. Consequently, the classification consistency from coarse to fine could be measured: 
\begin{equation}\label{eq12.8}
    {\rm JS}(s_j, s^{c}_{i}) = ({\rm KL}(avg, s_j) + {\rm KL}(avg, s^{c}_{i}))/2 \ , 
\end{equation}
\begin{equation}\label{eq12.9}
    avg = {\rm log}((s_j + s^{c}_{i})/2) \ , 
\end{equation}
and the consistency from fine to coarse can be measured with $s_{i}$ and $s^{f}_{j}$ as Eq. (\ref{eq12.8}).

\begin{figure}[t]
	\centering
    \includegraphics[width=0.45\linewidth]{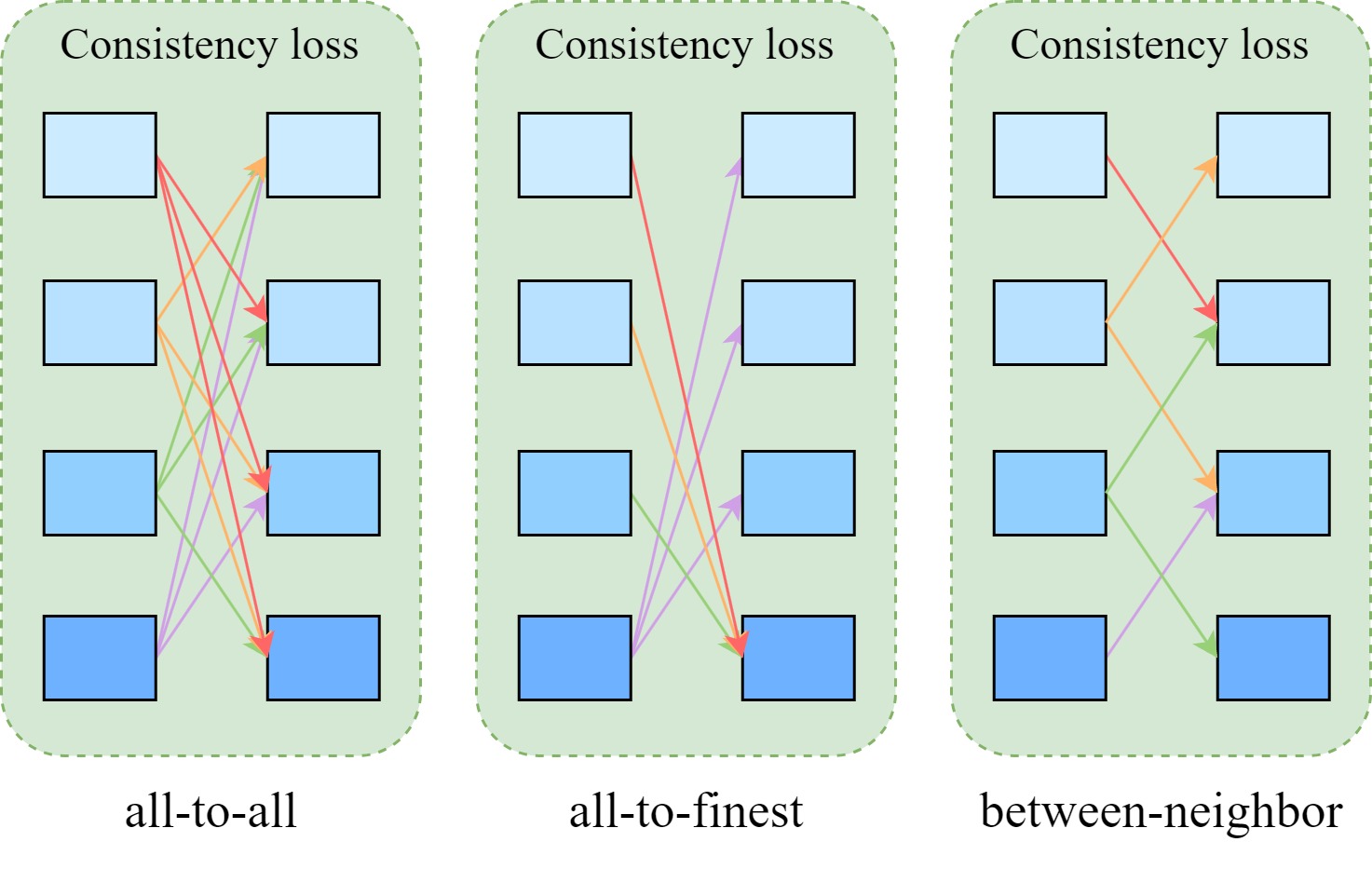}
	\caption{Three interaction strategies of consistency loss. All-to-all implies each level interacts with all other levels, all-to-finest implies each level only interacts with the finest level, between-neighbor implies each level only interacts with the neighbor in Tree Hierarchy.}
	\label{FIG:consistencyloss}
\end{figure}

A key problem arises when considering datasets whose depth of Tree Hierarchy is greater than 2 ($h \ge 3$): shall we consider the classification consistency across all hierarchies, only between neighboring hierarchies or between the finest level with others like strategies in Figure \ref{FIG:consistencyloss} ? 
CHBC strives to improve the consistency and to facilitate the diffusion of knowledge across all hierarchies. Based on this idea, we utilize all-to-all as the interaction strategy in CBC, encompassing broader contextual relations. Ablations also address this question in Section \ref{section:4.4}. 
Specifically, for the $l$-th hierarchy ($l \in [1, ..., h]$), we calculate the classification consistency between $l$-th level with all other levels, so that CHBC could unify all knowledge of Tree Hierarchy. 
After obtaining the coarse-to-fine distributions $s^{c}_{k < l}$ and the fine-to-coarse ones $s^{f}_{k > l}$ from the $k$-th ($k \ne l, k \in [1, ..., h]$) level, the combined distribution $\hat s_{l}$ is calculated as: 
\begin{equation}\label{eq14}
    \hat s_{l} = \sum_{k < l} s^{c}_k + \sum_{k > l} s^{f}_k \ , 
\end{equation}
$\hat s_l$ needs to be normalized to ensure the sum of $\hat s_{l}$ equal to 1. The consistency loss of all hierarchies is obtained by
\begin{equation}\label{eq15}
    \mathcal{L}_{con} = \sum_{l=1}^h {\rm JS}(s_l, \hat s_{l}) + {\rm JS}(s_{all}, \hat s_{all}) \ . 
\end{equation}

Eventually the loss function of CHBC consists of two parts: cross-entropy loss and consistency loss: 
\begin{equation}\label{eq16}
    \mathcal{L}_{cls} = \sum_{l=1}^{h} \mathcal{L}_{CE}({\rm FC}(F^p_l), y_l) + \mathcal{L}_{CE}({\rm FC}(F_{all}), y_h) \ , 
\end{equation}
\begin{equation}\label{eq17}
    \mathcal{L} = \mathcal{L}_{cls} + \mathcal{L}_{con} \ , 
\end{equation}
where $\mathcal{L}_{CE}$ denotes cross-entropy loss, $y_l$ means the ground truth of $l$-th hierarchy and $\rm FC(\cdot)$ is a fully connected layer.

\section{Experiments}\label{section:IV}

In this section, we evaluate CHBC on three FGVC datasets to demonstrate its effectiveness. Firstly, we briefly describe these datasets in Section \ref{section:4.1} and the implementation details in Section \ref{section:4.2}. Next, we analyze the results and compare our method with other sota methods to demonstrate the competitiveness of CHBC in Section \ref{section:4.3}. Finally, we perform ablations to validate the contributions of the proposed modules in Section \ref{section:4.4}.

\subsection{Datasets}\label{section:4.1}

\textbf{CUB-200-2011} \cite{CUB_200_2011} (CUB) is widely recognized as a benchmark in FGVC. It comprises 5,994 images for training and 5,794 images for testing. 
HSE \cite{chen_hse_mm18} extends the labels of CUB into Tree Hierarchy based on ornithological systematics and associates each image with four hierarchical labels (order, family, genus and species), organizing the 200 bird species into 122 genera, 37 families and 13 orders. The Tree Hierarchy in CUB is biologically informed, with varying numbers of subclasses for each non-leaf node. 

\textbf{FGVC-Aircraft} \cite{FGVC_Aircraft} (Air) is another frequently used dataset in FGVC, comprising 6,667 images for training and 3,333 images for testing. 
% It comprises 100 aircraft models, with 100 images per model, resulting in a total of 10,000 images—6,667 for training and 3,333 for testing. 
The Tree Hierarchy of Air has three levels, consisting of 30 makers, 70 families, and 100 models of aircraft. Similarly to CUB, the number of subclasses for non-leaf nodes varies. In addition, its Tree Hierarchy incorporates more human factors in its classification. 

\textbf{Stanford Cars} \cite{stanford_cars} (Cars) dataset contains 196 car models, which could be re-organized into 9 maker labels. It comprises 8,144 images for training and 8,041 images for testing. The Tree Hierarchy of Cars is unbalanced and incorporates human factors like Air. % More details are shown in Table \ref{dataset}. 

% \begin{table}[t]
% \captionsetup{width=\linewidth}
% \caption{The details of the FGVC datasets we utilize.}
% \label{dataset}
% % \begin{tabular*}{\dimexpr0.75\linewidth\relax}{@{}cccc@{}}
% \begin{tabular*}{\dimexpr0.38\linewidth\relax}{@{}cccc@{}}
% \toprule
% Dataset & Train & Test & Tree depth\\
% \midrule
% CUB-200-2011 & 5994 & 5794 & 4 \\
% FGVC-Aircraft & 6667 & 3333 & 3 \\
% Stanford Cars & 8144 & 8041 & 2 \\
% \bottomrule
% \end{tabular*}
% \end{table}

\subsection{Implementation details}\label{section:4.2}

To facilitate comparison with other methods that use ResNet-50 \cite{resnet50} as backbone \cite{Chen_hrn_cvpr22, chen_hse_mm18}, we set the first 41 convolutional layers of ResNet-50 pre-trained on ImageNet \cite{imagenet} to be the trunk net in CHBC, while setting the rest components to be MGE. Each MGE module contains two parallel last 9 convolutional layers of ResNet-50. The attention submodule is initialized with parameters pre-trained on ImageNet, while the predict submodule is initialized with parameters pre-trained on the specific datasets in Section \ref{section:4.1}. According to HSE \cite{chen_hse_mm18}, this initialization accelerates the convergence of models. 
The images are first scaled to 550$\times$550. During training, the images are transformed to 448×448 with random cropping, random horizontal flipping. AutoAugment \cite{autoaugment} is applied to Air and Cars. During testing, the images are cropped to 448×448 using center cropping without additional processing. The stochastic gradient descent with a momentum of 0.9 is used as the optimizer. The initial learning rate is set to 1e-2. An exponential decay schedule is applied for CUB and a cosine annealing scheduler is used for Air and Cars datasets. The enhancement factor in MGE is denoted as $\alpha = 0.4$, and the temperature coefficient in CBC is denoted as $T = 2$, their impacts are analyzed in Section \ref{section:4.4}. CHBC is implemented by PyTorch and trained on a Nvidia A40 GPU.

\subsection{Results and analysis}\label{section:4.3}

\textbf{Evaluation metrics. }
The classification difficulty varies from different levels, thus it is not reasonable to calculate the average accuracy of all hierarchies to assess the overall performance \cite{liu_chrf_eccv22}. We calculate the weighted average accuracy (wa\_acc) by summing accuracy $acc_m$ of $m$-th level with the number of categories $c_m$ as weights:
\begin{equation}
    {\rm wa\_acc} = \sum_{m=1}^h \frac{c_m}{\sum_{n=1}^h c_n} acc_{m} \ .
\end{equation}
To measure consistency, the metric Tree-based Consistency Rate (TCR) is proposed, denoting the rate of consistent and correct samples. TCR assesses the model's ability to implement containing relations within hierarchical categorization. Given Tree Hierarchy $\mathcal{TH}$, predicted labels $\{ \hat y_1, ..., \hat y_h \}$ and the finest ground truth $y_h$, TCR is calculated as: 
\begin{equation}\label{eq.tcr}
    {\rm TCR} = \frac{1}{N} \sum_{n=1}^N \mathbf{1} [ ( \{ \hat y_1, ..., \hat y_h \} \in \mathcal{TH} ) \land ( \hat y_h = y_h ) ] \ ,
\end{equation}
where $N$ is the number of samples, $\mathbf{1}(\cdot)$ is the indicator function.

\textbf{Compared with the sota methods. }
Since CHBC is the model that simultaneously outputs classification at multiple hierarchies, we mainly compare it with other methods that also output multi-level labels. Table \ref{multi_cub}, \ref{multi_air_cars} illustrate that CHBC achieves noticeable improvements over the baseline and sota methods on three datasets. 
Baseline-single categorizes each granularity with ResNet-50 independently. Baseline-multi uses ResNet-50 as the backbone and establishes branch net to predict multi-level labels simultaneously. With the incorporation of MGE and CBC modules, CHBC outperforms baseline-single and baseline-multi in all hierarchies. 
Among the four levels of baseline-multi on CUB, CHBC achieves the best in all levels, and the accuracy of CHBC improved by 3.1\% on the species level and 2.5\% on wa\_acc than baseline-multi. 
Compared with baseline-multi, CHBC reaches the optimum at all levels of Air and Cars, improving by 3.0\% and 2.1\% on the finest level in two datasets, and wa\_acc of CHBC scores the highest both on Air and Cars. CHBC also outperforms baseline-single, indicating that hierarchical knowledge could aid categorization. 
% In Section \ref{section:4.4}, compared with other sota models that also output multi-level labels, CHBC is also competitive on these datasets. % In the tables, red indicates the best result at each level and blue indicates the second-best result. 

\begin{table}[t]%[width=.8\linewidth,cols=5,pos=h]
\captionsetup{width=\linewidth}
\caption{Comparison with different approaches on CUB. The best and the second best accuracy (\%) and wa\_acc (\%) are respectively marked in \textbf{bold} and \underline{underline}.}
\label{multi_cub}
% \begin{tabular*}{\dimexpr0.88\linewidth\relax}{@{}ccccc@{}}
\begin{tabular*}{\dimexpr0.537\linewidth\relax}{@{}c | ccccc@{}}
\toprule
% \multirow{2}{*}{Method} & \multicolumn{5}{c}{Accuracy (\%)}\\
% \cmidrule{2-6}
Method  & Order & Family & Genus & Species & wa\_acc\\
\midrule
HMCN \cite{hmcn_icml18}    & 97.3 & 93.2 & - & 79.8 & 82.7\\
C-HMCNN \cite{C-HMCNN_NEURIPS2020}    & 98.5 & 94.6 & - & 81.6 & 84.4\\
FGN \cite{chang_fgn_cvpr21}    & 98.0 & 94.5 & 90.2 & 85.6 & 88.4\\
HRN \cite{Chen_hrn_cvpr22}    & 98.7 & 95.5 & - & 86.6 & 88.5\\
CAFL \cite{wang_cafl_mm23}    & \underline{99.1} & \textbf{96.3} & - & 87.5 & 89.4\\
HCSL \cite{liu_hcsl_pkdd24}    & \textbf{99.2} & \textbf{96.3} & - & 87.7 & 89.6\\
HSE \cite{chen_hse_mm18}     & 98.5 & 95.6 & 91.3 & \textbf{87.8} & \underline{90.1}\\
\midrule
Baseline-single & 98.5 & 95.2 & \underline{91.6} & 85.4 & 88.9\\
Baseline-multi & 98.6 & 94.3 & 90.2 & 84.7 & 87.9\\
Ours-CHBC      & \underline{99.1} & \underline{95.7} & \textbf{92.0} & \textbf{87.8} & \textbf{90.4}\\
\bottomrule
\end{tabular*}
\end{table}

\begin{table}[t]%[width=.7\linewidth,cols=4,pos=h]
\captionsetup{width=\linewidth}
\caption{Comparison with different approaches on Air and Cars. The best and the second best accuracy (\%) and wa\_acc (\%) are respectively marked in \textbf{bold} and \underline{underline}.}
\label{multi_air_cars}
% \begin{tabular*}{\dimexpr0.73\linewidth\relax}{@{}cccc@{}}
\begin{tabular*}{\dimexpr0.7\linewidth\relax}{@{}c | cccc | ccc@{}}
\toprule
% \multirow{3}{*}{Method} & \multicolumn{7}{c}{Accuracy (\%)}\\
% \cmidrule{2-8}
\multirow{2}{*}{Method}  & \multicolumn{4}{c|}{Air} & \multicolumn{3}{c}{Cars}\\
\cmidrule{2-5} \cmidrule{6-8}
  & Maker & Family & Model & wa\_acc & Maker & Model & wa\_acc\\
\midrule
HMCN \cite{hmcn_icml18}    & 96.1 & 92.6 & 87.2 & 90.4 & 95.2 & 88.7 & 89.0\\
% CAFL \cite{wang_cafl_mm23} & 96.4 & 94.3 & 89.0 \\
C-HMCNN \cite{C-HMCNN_NEURIPS2020}    & 97.5 & 95.4 & 91.7 & 93.9 & 96.8 & 90.6 & 90.9\\
HSE \cite{chen_hse_mm18}     & 97.8 & 95.0 & 92.3 & 94.1 & 97.2 & 93.6 & 93.7\\
FGN \cite{chang_fgn_cvpr21}     & 97.2 & 95.5 & 92.5 & 94.3 & 97.0 & 94.0 & 94.1\\
HRN \cite{Chen_hrn_cvpr22} & 97.5 & 95.8 & 92.6 & 94.5 & 97.4 & 94.0 & 94.1\\
HCSL \cite{liu_hcsl_pkdd24}    & \textbf{98.1} & \textbf{96.8} & \underline{93.0} & \underline{95.1} & \textbf{97.9} & \underline{94.6} & \underline{94.7}\\
\midrule
Baseline-single & 97.7 & 95.9 & 90.1 & 93.3 & 96.2 & 93.7 & 93.8\\
Baseline-multi & 96.8  & 95.5 & 90.6 & 93.2 & 96.4 & 93.2 & 93.3\\
Ours-CHBC & \underline{98.0} & \underline{96.5} & \textbf{93.6} & \textbf{95.3} & \underline{97.8} & \textbf{95.3}  & \textbf{95.4}\\  %92.3/
\bottomrule
\end{tabular*}
\end{table}

Compared with the sota multi-level models, CHBC demonstrates its effectiveness in all levels. 
The depth of CUB Tree Hierarchy is 4, while some approaches \cite{wang_cafl_mm23, Chen_hrn_cvpr22, liu_hcsl_pkdd24} only categorize the levels of order, family and species. We compare with HMCN \cite{hmcn_icml18} and C-HMCNN \cite{C-HMCNN_NEURIPS2020} by the reproduced results from HRN \cite{Chen_hrn_cvpr22}. HMCN \cite{hmcn_icml18} designs main flow and local flow to concatenate the global output with all local outputs for a consensual prediction. C-HMCNN \cite{C-HMCNN_NEURIPS2020} employs bottom and upper modules to produce level-specific outputs and imposes the hierarchy constraint. 

From Table \ref{multi_cub} and \ref{multi_air_cars}, we could observe that CHBC achieves the best accuracy at finer levels and wa\_acc on three datasets and performs its effectiveness at coarser levels, surpassing all methods on wa\_acc. 
FGN \cite{chang_fgn_cvpr21} and CAFL \cite{wang_cafl_mm23} divide the features outputted from backbone into equal-sized parts without any transformation. HCSL \cite{liu_hcsl_pkdd24} learns the fine-grained features depending on the coarse-grained ones, adjacency matrix and degree matrix in GCN. In contrast, CHBC extracts level-specific features from MGEs and highlights the discriminative features required at each hierarchy by CAM. 
Many models \cite{wang_cafl_mm23, Chen_hrn_cvpr22, chen_hse_mm18} attempt to constrain the classification consistency. HRN \cite{Chen_hrn_cvpr22} treats Tree Hierarchy as directed acyclic graph but calculates the consistency loss solely based on the directed edges, constraining the consistency in single direction. 
HSE \cite{chen_hse_mm18} also ignores the fine-to-coarse direction. 
CAFL \cite{wang_cafl_mm23} treats the classification probabilities as constants dependent on the number of subclasses when mapping coarse to fine. 
HMCN \cite{hmcn_icml18} neglects the consistency and implies all classes are independent. 
Contrarily, CHBC incorporates bidirectional consistency constraints, enabling knowledge to be transferred from two directions, and the classification accuracy of coarse levels can also be improved because of the constraints from other levels. Additionally, CHBC imposes consistency constraints that align with human's classification intuition. When the classification at coarser level is confidently determined, CHBC tends to select a fine-grained label from the subclasses of the identified superclass. 

Table \ref{TOP35-TCR} shows the comparison of Top-3/5 wa\_acc and TCR. Compared with the methods we retrain, CHBC still achieves best on three datasets. CBC contributes to the improvement of Top-3/5 wa\_acc by boosting the probabilities of subclasses under correct superclass, which could delivers more accurate categories for indistinguishable images. TCR of CHBC also reaches the highest, indicating the effectiveness of bidirectional consistency constraints.

\begin{table}[t]
\caption{Comparison of the Top-3/5 wa\_acc (\%) and TCR (\%) on CUB, Air and Cars. The best and the second best results are respectively marked in \textbf{bold} and \underline{underline}.}
\label{TOP35-TCR}
\begin{tabular*}{\dimexpr0.631\linewidth\relax}{@{}ccccccc@{}}
\toprule
\multirow{2}{*}{Method} & \multicolumn{2}{c}{CUB} & \multicolumn{2}{c}{Air} & \multicolumn{2}{c}{Cars}\\
\cmidrule{2-7}
 & Top-3/5 & TCR & Top-3/5 & TCR & Top-3/5 & TCR \\
\midrule
HSE \cite{chen_hse_mm18} & \underline{96.5}/\underline{97.8} & \underline{83.5}    & \underline{97.4}/\underline{97.7} & 91.4    & \underline{98.6}/\underline{99.1} & 92.4 \\
FGN \cite{chang_fgn_cvpr21} & 95.8/97.2 & 82.6    & 97.2/\underline{97.7} & \underline{91.5}    & \underline{98.6}/\underline{99.1} & \underline{92.6} \\
\midrule
Baseline-multi & 95.1/96.7 & 81.6    & 96.6/97.0 & 89.0    & 97.9/98.4 & 92.1 \\
Ours-CHBC & \textbf{96.7}/\textbf{97.9} & \textbf{85.0}    & \textbf{98.1}/\textbf{98.6} & \textbf{92.5}    & \textbf{99.1}/\textbf{99.6} & \textbf{94.3} \\
\bottomrule
\end{tabular*}
\end{table}

\begin{table}[t]%[width=.9\linewidth,cols=4,pos=h]
\caption{Comparison of CHBC with the sota methods outputing the finest labels only on CUB, Air and Cars. The best and the second best accuracy (\%) are respectively marked in \textbf{bold} and \underline{underline}.}
\label{single_sota}
% \begin{tabular*}{\dimexpr0.82\linewidth\relax}{@{}ccccc@{}}
\begin{tabular*}{\dimexpr0.41\linewidth\relax}{@{}ccccc@{}}
\toprule
\multirow{2}{*}{Method} & \multirow{2}{*}{Backbone} & \multicolumn{3}{c}{Accuracy}\\
\cmidrule{3-5}
 & & CUB & Air & Cars \\
\midrule
Cross-X \cite{luo_crossx_iccv19} &  ResNet-50 & 87.7 & 92.6 & 94.6 \\
MC-Loss \cite{chang_mcloss_tip20}  &  ResNet-50 & 87.3 & 92.6 & 93.7 \\
CIN \cite{gao_cin_aaai20}  &  ResNet-50 & 87.5 & 92.6 & 94.1 \\
PA-CNN \cite{zheng_pacnn_tip20} &  VGG-19 & 87.5 & 91.0 & 93.3 \\
API-Net \cite{zhuang_apinet_aaai20}  &  ResNet-50 & 87.7 & 93.0 & \underline{94.8} \\
AKEN \cite{hu_aken_tcsvt21} &  VGG-19 & 87.1 & 94.0 & 93.9 \\
SFFF \cite{wang_sfff_tcsvt23} &  ResNet-50 & 85.4 & \underline{93.1} & 94.4 \\
% AP-CNN & TIP 2021 & ResNet-50 & 87.2 \\
SAC \cite{do_sac_cai24} &  ResNet-50 & \textbf{88.3} & 92.1 & - \\
\midrule
Ours-CHBC & ResNet-50 & \underline{87.8} & \textbf{93.6} & \textbf{95.3} \\
\bottomrule
\end{tabular*}
\end{table}

Table \ref{single_sota} compares CHBC with the sota models that only produce the finest-grained classification. Since the backbone of CHBC is ResNet-50, we primarily compare with the methods that also utilize ResNet-50 as their backbone. 
We retain the accuracy of the finest hierarchy in CHBC to compare with other single-level approaches. CHBC demonstrates competitive performance compared to these sota single-level models and achieves the best on Air and Cars datasets. This also validates that CHBC effectively transfers knowledge across levels, which could assist in improving the classification accuracy.

\begin{figure*}[t]
	\centering
	\includegraphics[width=\textwidth]{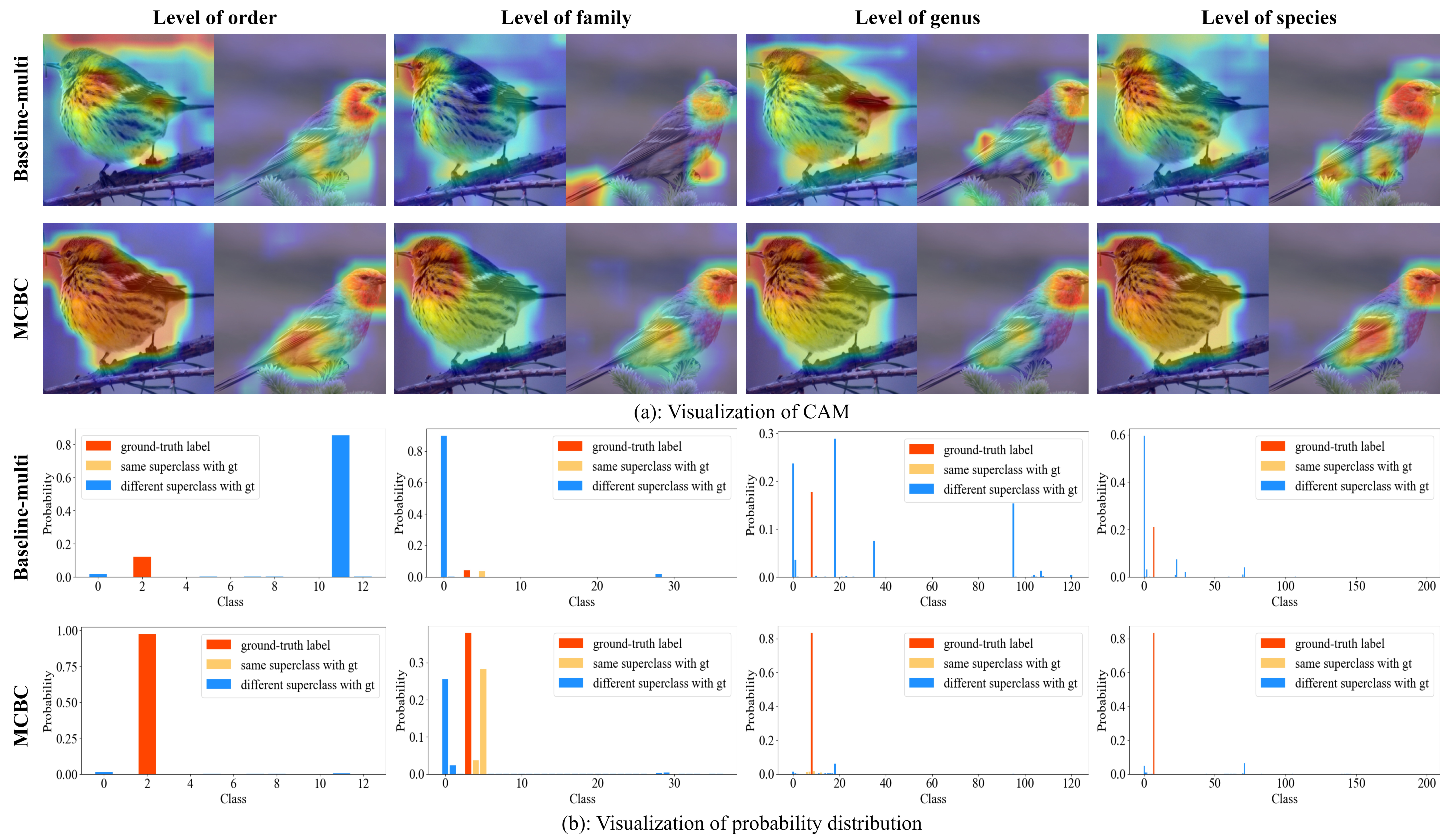}
	\caption{(a) is the visualization of CAM, and (b) is the visualization of probability distributions on CUB. }
	\label{FIG:cub_attition_probability}
\end{figure*}

\textbf{Visualization and Analysis. }
CHBC improves classification accuracy by constraining the consistency and enhancing the consistent distinct features across different hierarchies. 
The consistency loss in CBC encourages that subclasses within the same predicted superclass have higher probability if this superclass has high predicted probability. For instance, the hierarchical ground truths of an image are [\textit{Albatross, Laysan Albatross}], and CHBC predicts the coarse label as \textit{Albatross}, then CBC tends to increasing the probability of \textit{Laysan Albatross} while also increasing the probability of \textit{Black-footed Albatross} because of the same superclass they belong to. Thereby, CBC could improve the Top-3 and Top-5 accuracy when an image is indistinguishable among subclasses.
We visualize the classification probabilities of each level in CHBC and compare them with baseline-multi, as shown in Figure \ref{FIG:cub_attition_probability} (b). The red denotes the probability of ground truth (gt), the yellow represents the subclasses that belong to the same superclass with gt and the blue means the classes whose coarse label is different with gt. 
It can be observed that, CHBC predicts properly on all four levels and corrects the misclassification of baseline-multi. The probabilities of yellow classes increase, and the other blues tend to keeping low values, which could improve the accuracy of Top-$n$ ($n \ge 1$). In Figure \ref{FIG:cub_attition_probability} (b), this phenomenon is reflected in all levels and the misclassifications of baseline-multi are corrected. 
In the level of family, CHBC predicts the gt successfully and increases two yellow bars dramatically. But baseline-multi predicts a great error which the predicted label belongs to different superclass with gt. % It demonstrates that CBC performed its intended function. 
This can be interpreted as humans identifying coarse labels and focusing more attention on the fine labels contained in this coarse label. Although this phenomenon will lead to increased cross-entropy loss, it could be viewed as an equilibrium reached after the competition between cross-entropy loss and consistency loss. 

\begin{figure}[t]
	\centering
    \includegraphics[width=0.5\linewidth]{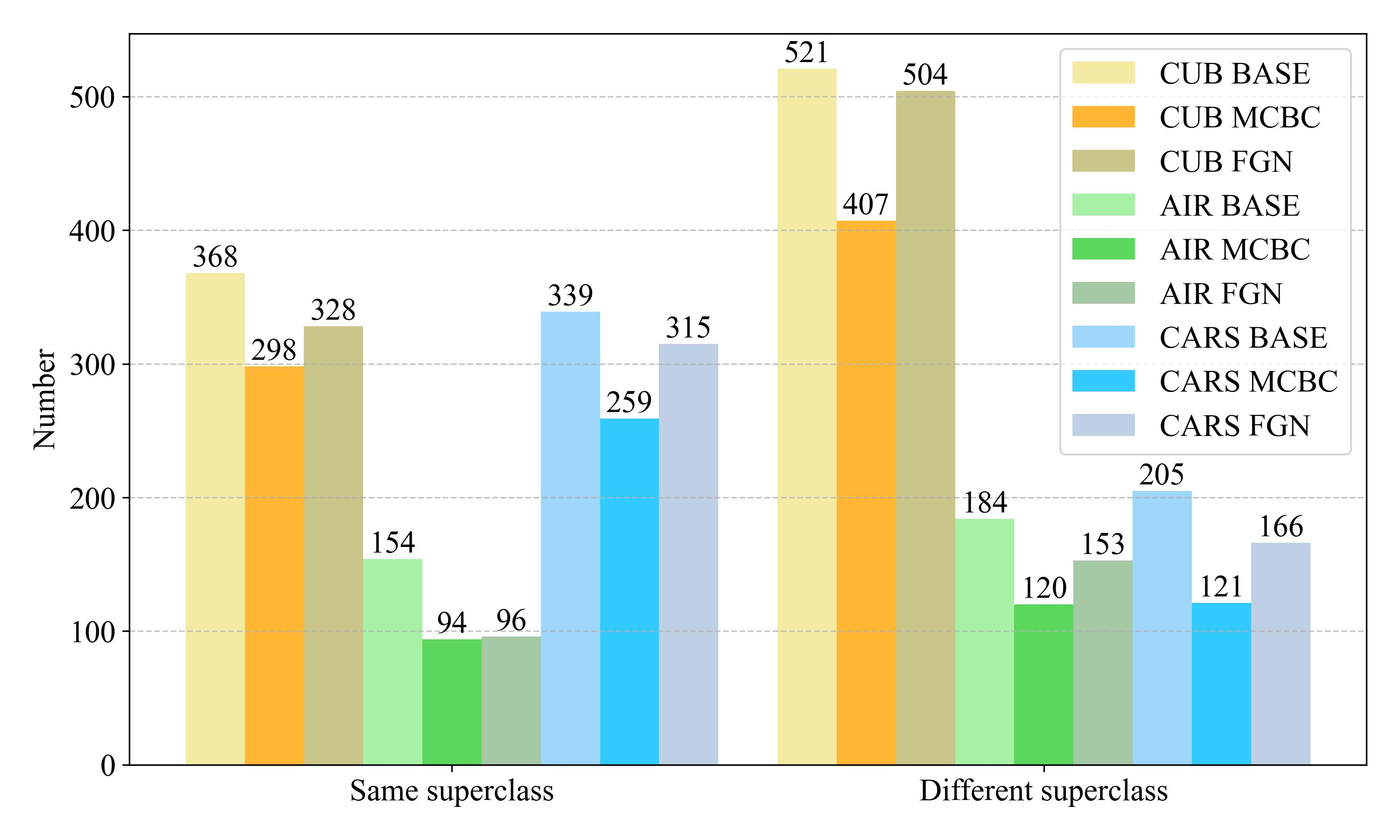}
	\caption{The histogram of the number of misclassified samples, which records the number of predicted labels and ground truth belonging to the same superclass or not when misclassified.}
	\label{FIG:sameornot}
\end{figure}
It also aligns with our idea: if model makes a fine-grained classification error, this error should not be too conspicuous and the fine prediction should be as close to the gt as possible. In other words, if the fine gt of an image is \textit{A} but model misclassifies as \textit{B} (\textit{A} $\ne$ \textit{B}) in the same level, we hope that \textit{A} and \textit{B} belong to the same superclass. In Figure \ref{FIG:sameornot}, the left represents the cases in which the predicted labels and the gt belong to the same superclass when misclassified, while the right represents the cases they belong to different superclasses. 
Compared with baseline-multi and FGN, the number of misclassification under different superclasses in CHBC decreases more than under the same superclass. CHBC increases the accuracy while reducing more misclassification under different superclasses significantly, further confirming the effectiveness of our model in keeping misclassified labels as close as possible to the gt labels.

The CAMs of CHBC for image classification are visualized and compared with those of baseline-multi in Figure \ref{FIG:cub_attition_probability} (a), the first row shows the different birds' visualization of baseline-multi in multiple levels, while the second row depicts the results of CHBC. During the forward propagation, the CAMs of baseline-multi focus on more scattered regions and include more background information unrelated to objects. In contrast, the CAMs of CHBC are more concentrated on objects at the finer granularity and keep the consistency of interesting regions. Between different kinds of bird, CHBC is concerned with distinguish features. CHBC pays more attention to the head and breast in the left, and focuses on the head, wings and belly in the right.

\subsection{Ablation study}\label{section:4.4}

We conduct ablations on CHBC to investigate the effectiveness of the proposed modules and the strategies in these modules.
The results of baseline-multi, representing the results without any strategies in any modules, are shown in the first row of Table \ref{ablation_cub_mge}, \ref{ablation_cub_cbc} and \ref{ablation_cub1}. 

Table \ref{ablation_cub_mge} reports the results of various strategies implemented in MGE. Rows 2$\sim$4 illustrate three types of enhancing matrices. The strategy \textit{AddAll}, referring to the ideas of residual connection, involves summing the matrices of each level with those of all preceding levels, \textit{AddPre} entails adding the matrices of each level with the previous level only, and \textit{MatOrth} represents performing orthogonal decomposition between hierarchical levels, which is employed by CHBC as Figure \ref{FIG:details_in_module} (a). Notably, the improvement achieved by \textit{MatOrth} is more pronounced for the finest labels. Compared with others, \textit{MatOrth} calculates and strengthens the orthogonal components between matrices at different levels. 
Rows 5$\sim$7 show the exploration of attention and features extracted from which levels. $A^{pre}F^{1-th}$ involves enhancing the current level with attention from the previous level and features from the first level; $A^{pre}F^{pre}$ involves using the attention and features both from the previous level; and $A^{1-th}F^{pre}$ involves using attention from the first level and features from the previous level. CHBC employs $A^{pre}F^{1-th}$, which yields the most substantial improvement in performance. 
Enhancement between neighboring levels is the most effective approach to extract discriminative information. However, applying the same strategy to attention and features would reduce diversity of our model. 

\begin{table}[t]%[width=.75\linewidth,cols=6,pos=h]
\captionsetup{width=\linewidth}
\caption{Ablations of different strategies in MGE on CUB. Rows 2$\sim$4 show three types of enhancement in MGE like Figure \ref{FIG:details_in_module} (a) and rows 5$\sim$7 show three strategies for which levels to be utilized in enhancement. }
\label{ablation_cub_mge}
% \begin{tabular*}{\dimexpr0.78\linewidth\relax}{@{}ccccc@{}}
\begin{tabular*}{\dimexpr0.48\linewidth\relax}{@{}cccccc@{}}
\toprule
\multirow{2}{*}{MGE} & \multicolumn{5}{c}{Accuracy (\%)} \\
\cmidrule{2-6}
   & Order & Family & Genus & Species & wa\_acc \\
\midrule
 Base & 98.6 & 94.3 & 90.2 & 84.7 & 88.0 \\
\midrule
 \textit{AddAll} & 98.8 & 95.4 & 91.4 & 86.6 & 89.5 \\
 \textit{AddPre} & 98.7 & 94.8 & 91.2 & 86.8 & 89.5 \\
 \textit{MatOrth} & \textbf{99.1} & \textbf{95.7} & \textbf{92.0} & \textbf{87.8} & \textbf{90.4} \\
\midrule
$A^{1-th}F^{pre}$ & 98.7 & 95.0 & 91.4 & 87.5 & 89.9 \\
$A^{pre}F^{pre}$ & 98.5 & 94.9 & 91.6 & 87.3 & 89.9 \\
$A^{pre}F^{1-th}$ & \textbf{99.1} & \textbf{95.7} & \textbf{92.0} & \textbf{87.8} & \textbf{90.4} \\
\bottomrule
\end{tabular*}
\end{table}

\begin{table}[t]%[width=.75\linewidth,cols=6,pos=h]
\captionsetup{width=\linewidth}
\caption{Ablations of different strategies in CBC on CUB. Rows 2$\sim$4 show ablation of three distance measurements and rows 5$\sim$7 show three interaction strategies in CBC like Figure \ref{FIG:consistencyloss}. }
\label{ablation_cub_cbc}
% \begin{tabular*}{\dimexpr0.78\linewidth\relax}{@{}ccccc@{}}
\begin{tabular*}{\dimexpr0.483\linewidth\relax}{@{}cccccc@{}}
\toprule
\multirow{2}{*}{CBC} & \multicolumn{5}{c}{Accuracy (\%)} \\
\cmidrule{2-6}
   & Order & Family & Genus & Species & wa\_acc\\
\midrule
 Base & 98.6 & 94.3 & 90.2 & 84.7 & 88.0 \\
\midrule
 KL & 98.8 & 95.4 & 91.5 & 87.2 & 89.8 \\
 EMD & 98.7 & 94.6 & 90.2 & 84.1 & 87.7 \\
 JS & \textbf{99.1} & \textbf{95.7} & \textbf{92.0} & \textbf{87.8} & \textbf{90.4} \\
\midrule
 \textit{Neighbor} & 98.9 & 94.7 & 91.0 & 86.5 & 89.2 \\
 \textit{Finest} & 98.7 & 94.8 & 91.2 & 86.2 & 89.1 \\
 \textit{All} & \textbf{99.1} & \textbf{95.7} & \textbf{92.0} & \textbf{87.8} & \textbf{90.4} \\
\bottomrule
\end{tabular*}
\end{table}

Table \ref{ablation_cub_cbc} shows different distance measurements and interaction strategies in CBC. 
Rows 2$\sim$4 illustrate three functions to calculate the distance between distributions. 
KL demands that the coarse or fine distribution is defined as a reference so that another distribution could be aligned with it. 
However, it is unjustifiable to simply specify the distribution of coarse, fine or current level as the reference. Therefore, we utilize the JS divergence, which overcomes this problem. 
Earth-Mover Distance (EMD), also known as Wasserstein Distance, is another function for measuring the distance between distributions. Rows 2$\sim$4 in Table \ref{ablation_cub_cbc} reveals that JS divergence yields the best performance. This conclusion also confirms that setting the probability distribution of a specific level (coarse or fine) as the reference in CBC is inappropriate, and it makes more sense to consider the impact of both directions. 
The last three rows demonstrate the improvements made by different interaction strategies used in CBC. \textit{Neighbor}, \textit{Finest} and \textit{All} represent between-neighbor, all-to-finest and all-to-all respectively depicted in Figure \ref{FIG:consistencyloss}. 
All-to-all involves the probability distribution of each level influencing all other levels, thereby ensuring consistency of classification results across all hierarchies. It facilitates the transfer of knowledge across hierarchies and is not limited to between only two levels, further guiding the harmonization of features learned from MGEs. 
In contrast, all-to-finest only addresses the consistency between the finest level and a single coarse level, and between-neighbor constrains no consistency between the nonadjacent levels. The experiments further demonstrate that the accuracy of all-to-all is higher than other strategies. 

\begin{table}[t]%[width=.75\linewidth,cols=6,pos=h]
\captionsetup{width=\linewidth}
\caption{Ablation experiments of the proposed modules on CUB.}
\label{ablation_cub1}
% \begin{tabular*}{\dimexpr0.85\linewidth\relax}{@{}cccccc@{}}
\begin{tabular*}{\dimexpr0.512\linewidth\relax}{@{}ccccccc@{}}
\toprule
\multicolumn{2}{c}{Method} & \multicolumn{5}{c}{Accuracy (\%)}\\
\cmidrule{3-7}
MGE & CBC & Order & Family & Genus & Species & wa\_acc \\
\midrule
 - & - & 98.6 & 94.3 & 90.2 & 84.7 & 88.0 \\
\checkmark & - & 98.5 & 94.7 & 90.8 & 86.2 & 89.0 \\
 - & \checkmark & 98.7 & \textbf{95.8} & 91.7 & 87.0 & 89.8 \\
 \checkmark & \checkmark & \textbf{99.1} & 95.7 & \textbf{92.0} & \textbf{87.8} & \textbf{90.4} \\
\bottomrule
\end{tabular*}
\end{table}

Table \ref{ablation_cub1} reports the ablation of the proposed modules and \checkmark denotes the application of this module. When only MGE is applied, the finest-grained accuracy improves by 1.5\% and wa\_acc improves by 1.0\%. Applying only CBC further improves the finest-grained accuracy by 2.3\% compared to baseline-multi, which could be attributed to its ability to constrain the consistency. 
Compared to the baseline-multi, CHBC achieves an overall improvement of 3.1\% at the species level and improves by 2.4\% at wa\_acc. 
These experiments demonstrate the noticeable contributions of MGE and CBC in improving accuracy.

\begin{figure}[t]
    \begin{minipage}{0.45\linewidth}
        \vspace{1pt}
        % \centerline{\includegraphics[width=1.1\linewidth]{figs/ablation_T.jpg}}
        \centerline{\includegraphics[width=0.75\linewidth]{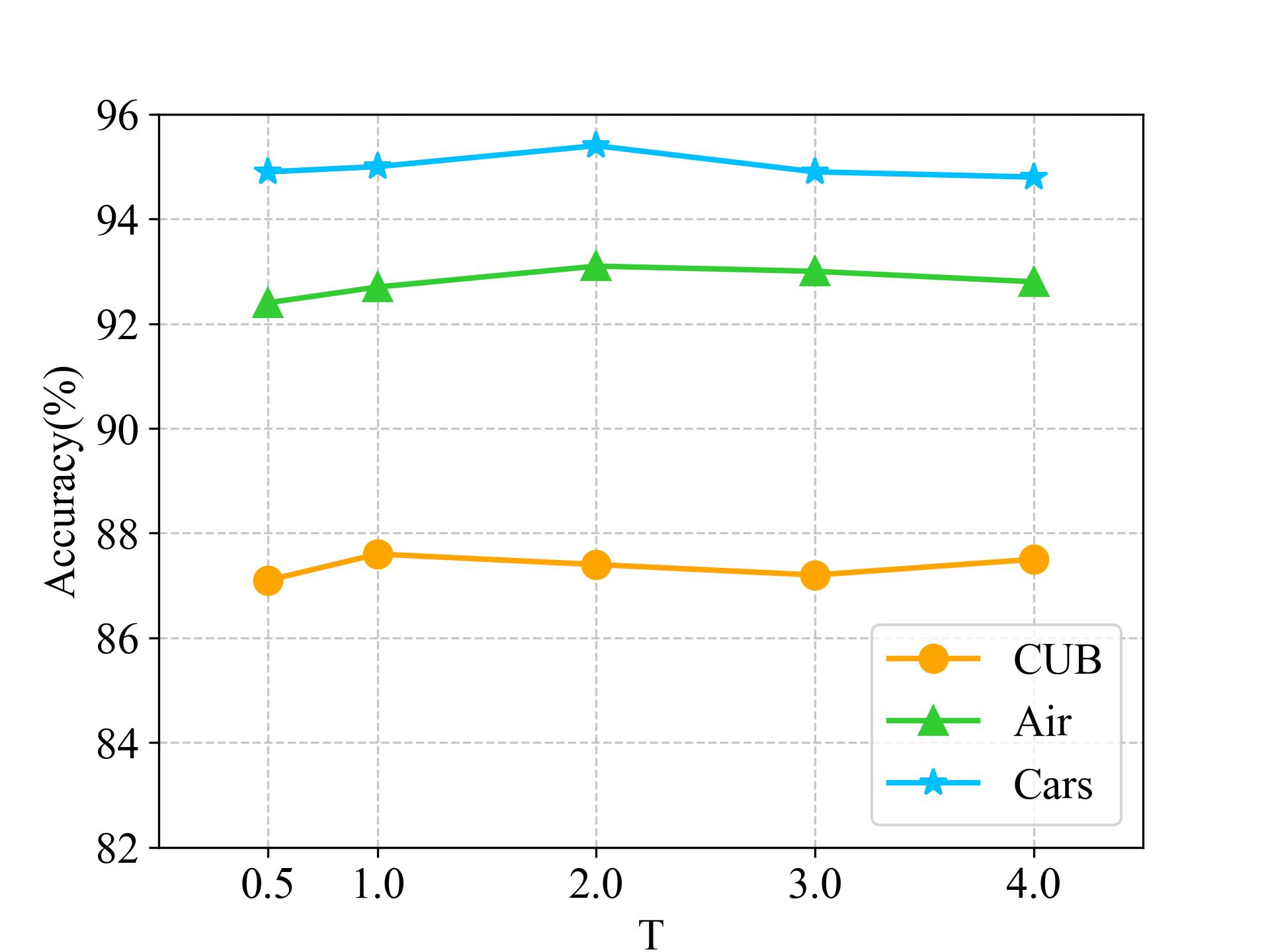}}
        \centerline{T with different values}
    \end{minipage}
    \begin{minipage}{0.45\linewidth}
        \vspace{1pt}
        % \centerline{\includegraphics[width=1.1\linewidth]{figs/ablation_alpha.jpg}}
        \centerline{\includegraphics[width=0.75\linewidth]{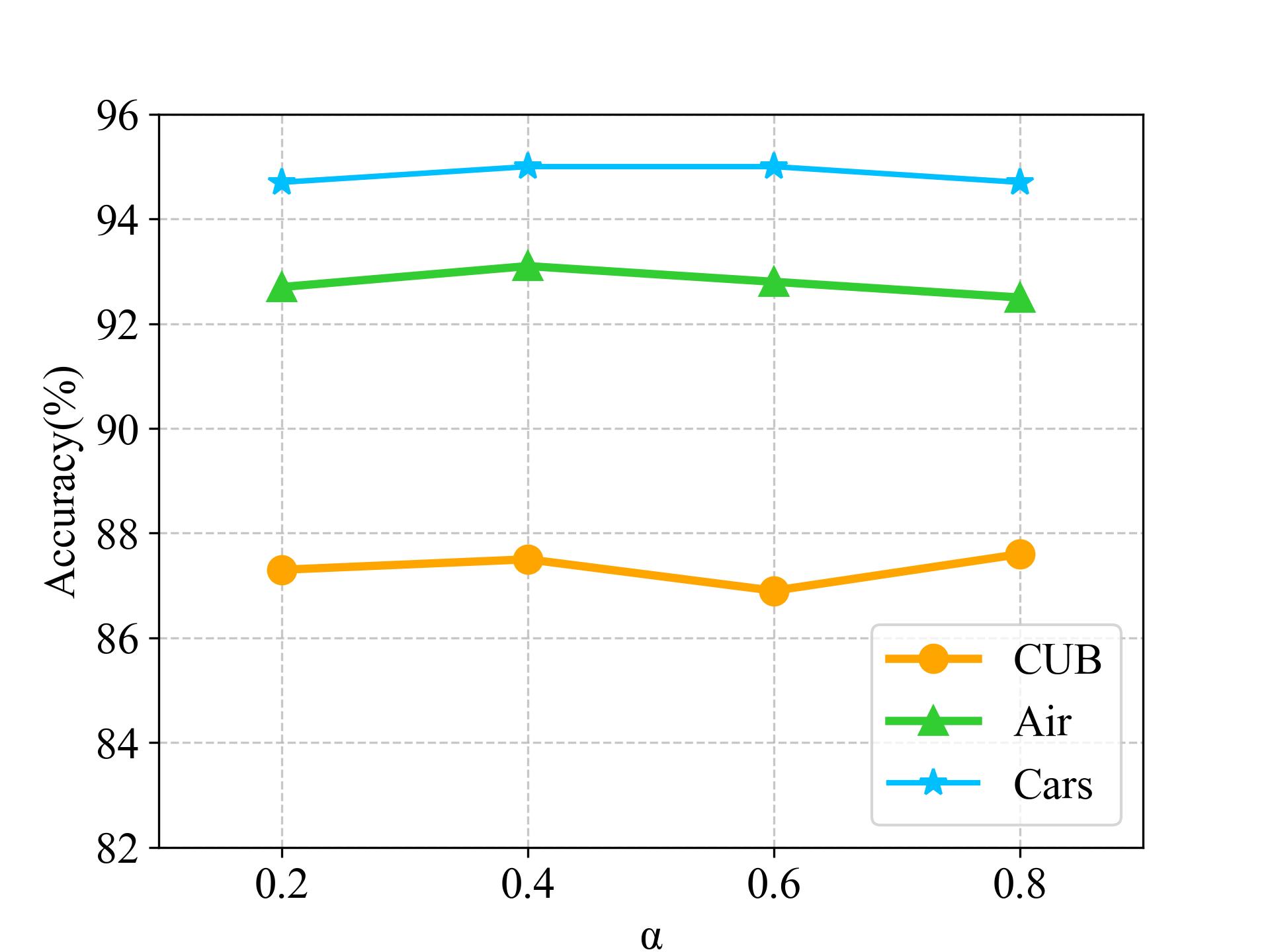}}
        \centerline{$\alpha$ with different values}
    \end{minipage}
	\caption{The effect of T and $\alpha$ with different values on CUB, Air and Cars.}
	\label{FIG:albation}
\end{figure}

We visualize the effects of two important hyperparameters as shown in Figure \ref{FIG:albation}. $T$ is the temperature coefficient that modulates the probability distribution. According to HSE \cite{chen_hse_mm18}, a higher temperature coefficient leads to a smoother probability distribution. $\alpha$ is the enhancement coefficient in MGE. A larger $\alpha$ results in a more pronounced enhancement of features in the fine-grained level. However, the model's efforts at the fine-grained level may increasingly focus on background information that is unrelated to the object at the coarse-grained level.

\section{Conclusion}\label{section:V}

In this paper, we study the FGVC problem with hierarchical labels, aiming at multi-label classification without introducing additional annotations, but the inherent hierarchical knowledge. 
We propose CHBC and partition FGVC into two components for resolution. To extract discriminative features at each hierarchy, we utilize the MGE module to decompose and enhance attention masks and features across different levels. To ensure consistency in classification results across hierarchies, we introduce the CBC module, which constrains consistency bidirectionally. Finally, we experiment on three datasets with hierarchical labels. 
Experiments demonstrate that CHBC effectively exploits inter-level consistency and improves classification accuracy.

% \bibliographystyle{cas-model2-names}
\bibliographystyle{elsarticle-num}

\bibliography{cas-refs}

\end{document}